\documentclass[10pt, letter, twocolumn]{IEEEtran}

\usepackage{graphicx}
\usepackage{epsfig}
\usepackage{algorithm}
\usepackage{algorithmic}
\usepackage{ifmtarg}
\usepackage[font=small,labelfont=bf]{caption}
\usepackage{placeins}
\usepackage{footnote}
\usepackage{cite}
\usepackage{amssymb}
\usepackage{amsthm}
\usepackage[fleqn]{amsmath}
\usepackage{amsmath}
\usepackage{amsfonts}
\usepackage{epstopdf}
\usepackage{enumitem}
\usepackage{subfigure}
\usepackage{multirow}
\usepackage{rotating}
\usepackage{amsmath}

\usepackage[utf8]{inputenc}
\usepackage[english]{babel}

\usepackage{todonotes}
\begin{document}

\title{Autonomous UAV Navigation: A DDPG-based Deep Reinforcement Learning Approach}

\author{\IEEEauthorblockN{Omar Bouhamed$^{1}$, Hakim Ghazzai$^1$, Hichem Besbes$^2$ and Yehia Massoud$^1$}\\
\IEEEauthorblockA{\small $^1$School of Systems \& Enterprises, Stevens Institute of Technology, Hoboken, NJ, USA \\
%Email: \{obouhame, hghazzai, ymassoud\}@stevens.edu
$^2$University of Carthage, Higher School of Communications of Tunis, Tunisia\\
%Email: \{omar.bouhamed, hichem.besbes\}@supcom.tn\\
}\vspace{-0.3cm}
{\thanks {\hrule

\vspace{0.1cm} \indent This paper is accepted for publication in IEEE International Symposium on Circuits and Systems (ISCAS'20), Seville, Spain, Oct. 2020.

2020 IEEE. Personal use of this material is permitted. Permission from IEEE must be obtained for all other uses, in any current or future media,including reprinting/republishing this material for advertising or promotional purposes, creating new collective works, for resale or redistribution to servers or lists, or reuse of any copyrighted component of this work in other works.
}}
}

\maketitle
\thispagestyle{empty}
\pagestyle{empty}

\begin{abstract}
In this paper, we propose an autonomous UAV path planning framework using deep reinforcement learning approach. The objective is to employ a self-trained UAV as a flying mobile unit to reach spatially distributed moving or static targets in a given three dimensional urban area. In this approach, a Deep Deterministic Policy Gradient (DDPG) with continuous action space is designed to train the UAV to navigate through or over the obstacles to reach its assigned target. A customized reward function is developed to minimize the distance separating the UAV and its destination while penalizing collisions. Numerical simulations investigate the behavior of the UAV in learning the environment and autonomously determining trajectories for different selected scenarios.\vspace{-0.2cm}
\end{abstract}

\begin{IEEEkeywords}
Autonomous navigation, deep reinforcement learning, obstacle avoidance, unmanned aerial vehicle.\vspace{-0.1cm}
\end{IEEEkeywords}

\vspace{-.2cm}
\section{Introduction}\label{Introduction}
Smart cities are witnessing a rapid development to provide satisfactory quality of life to its citizens~\cite{7539244}. The establishment of such cities requires the integration and use of novel and emerging technologies. In this context, unmanned areal vehicles (UAV), \textit{aka} drones, are continuously proving their efficiency in leveraging multiple services in several fields, such as good delivery and traffic monitoring (e.g. Amazon is starting to use UAVs to deliver packages to customers). UAVs are easy to deploy with a three dimensional (3D) mobility as well as a flexibility in performing difficult and remotely located tasks while providing bird-eye view~\cite{8613833,8731861}. Path planning remains one of key challenges that need to be solved to improve UAV navigation especially in urban areas. The core idea is to devise optimal or near-optimal collision-free path planning solutions to guide UAVs to reach a given target, while taking into consideration the environment and obstacle constraints in the area of interest.

In recent studies, such as~\cite{8359869}, the authors adopted the ant colony optimization algorithm to determine routes for UAVs while considering obstacle avoidance for modern air defence syste. In~\cite{8833455}, a combination of grey wolf optimization and fruit fly optimization algorithms is proposed for the path planning of UAV in oilfield environment. In~\cite{7978637,8747351,xian}, the UAV path planning problems were modeled as mixed integer linear programs (MILP) problem. Also, in~\cite{7829099}, a 3D path planning method for multi-UAVs system or single UAV is proposed to find a safe and collision-free trajectory in an environment containing obstacles. However, most of the solutions are based on MILP which are computationally complex or evolutionary algorithms, which do not necessarily reach near-optimal solutions. Moreover, the existing approaches remain centralized where a central node, e.g. a control center runs the algorithm and provides to the UAV its path plan. Centralized approaches restrain the system and limit its capabilities to deal with real-time problems. They impose a certain level of dependency and cost additional communication overhead between the central node and the flying unit. Hence, artificial intelligence (AI), precisely, reinforcement learning (RL) come out as a new research tendency that can grant the flying units sufficient intelligence to make local decisions to accomplish necessary tasks. In~\cite{8443226} and \cite{leme}, the authors presented a Q-learning algorithm to solve the autonomous navigation problem of UAVs. Q-learning was also employed to establish paths while avoiding obstacles in~\cite{8742915}. However, the authors used discrete actions (i.e. the environment is modeled as a grid world with limited UAV action space, degree of freedom). which may reduce the UAV efficiency while dealing with real-world environment, where the flying units operate according to a continuous action space.
\begin{figure}[t!]
  \centerline{\includegraphics[width=7cm]{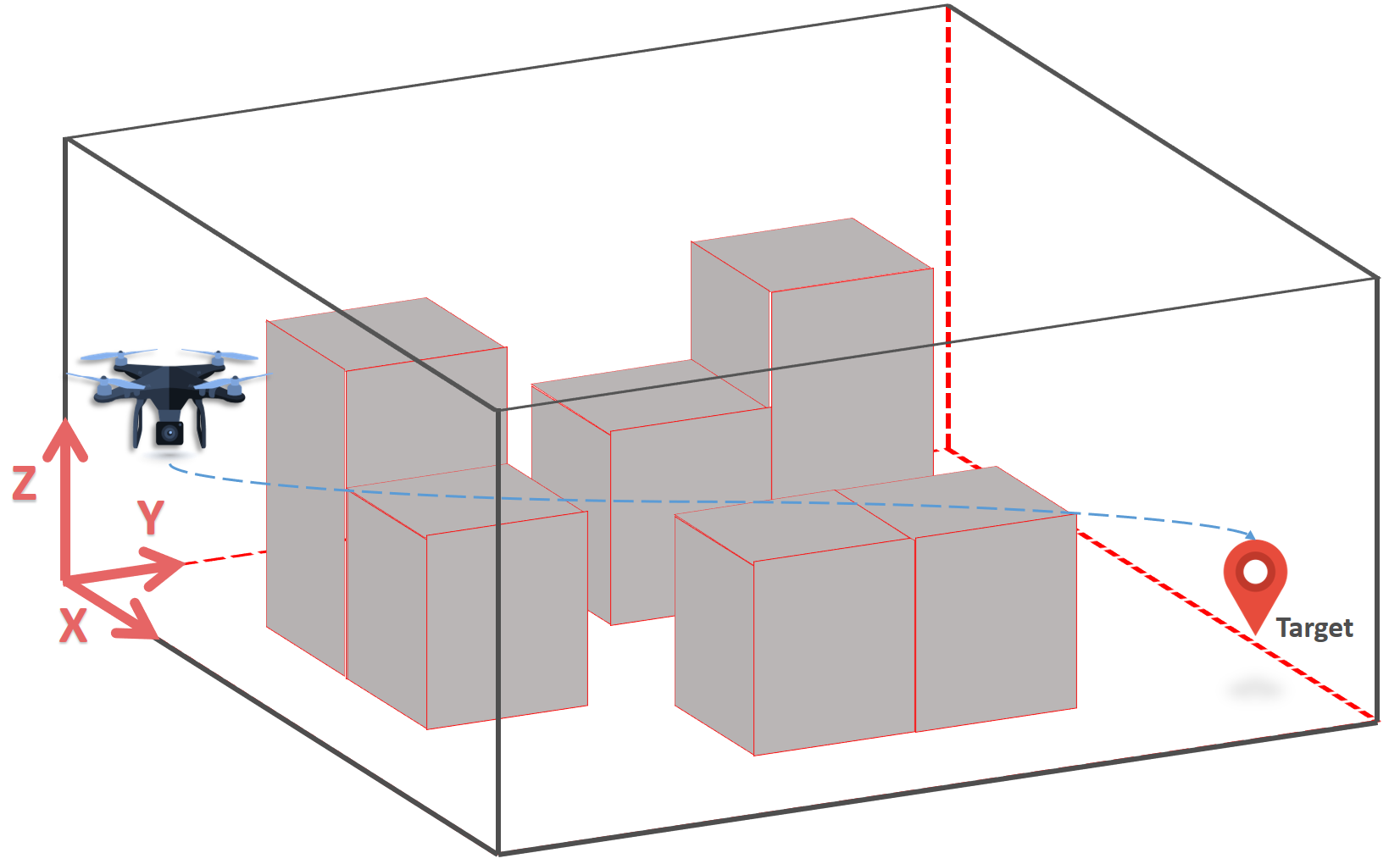}}\vspace{-0.2cm}
   \caption{Illustration of the autonomous obstacle-aware UAV navigation in an urban environment.}\label{system}\vspace{-0.5cm}
\end{figure}

Unlike existing RL-based solutions which are usually operating on a discretized environment, the proposed framework aims to provide UAV autonomous navigation with continuous action space to reach fixed or moving targets dispersed within a 3D space area while considering the UAV safety. A deep deterministic gradient decent (DDPG)-based approach is modeled with the objective to allow an UAV determine the best course to accomplish its missions safely, i.e. obstacle avoidance. A reward function is designed to guide the UAV toward its destination while penalizing any crash. During the training phase, we adopt a transfer learning approach to train the UAV how to reach its destination in a free-space environment (i.e., source task). Then, the learned model is fed to other models (i.e., new task) dedicated to different environments with specific obstacles' locations so that the UAV can learn how to avoid obstacles to navigate to the destination. During the prediction phase, it determines the path within the training environment by figuring out which route to take to reach any randomly generated static or dynamic destination from any arbitrary starting position. In the simulations, we investigate the behavior of the autonomous UAVs for different scenarios including obstacle-free and urban environments. It is shown that the UAV smartly selects paths to reach its target while avoiding obstacles either by crossing over or deviating them.

\begin{figure}[t!]
  \centerline{\includegraphics[width=8cm]{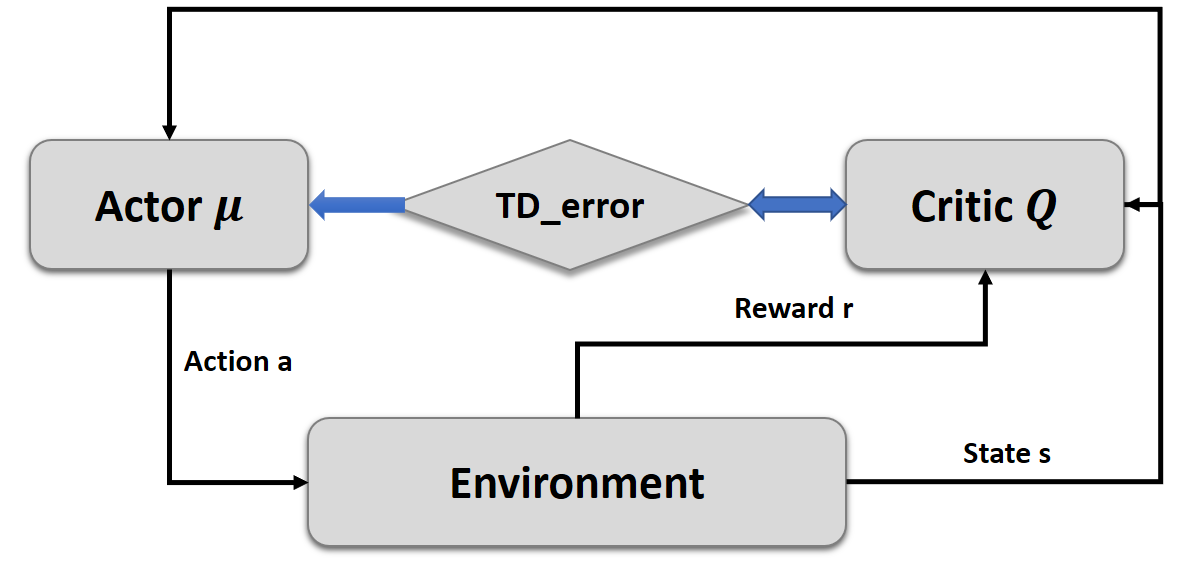}}\vspace{-0.1cm}
   \caption{Architecture of the actor-critic learning model.}\label{alg_fig}\vspace{-0.4cm}
\end{figure}
\section{Problem Description}\label{SystemModel}
\textcolor{black}{In this section, we present the system model and describe the actions that can be taken by the UAV to enable its autonomous navigation. }
\vspace{-0.3cm}
\subsection{System Model}
Autonomous navigation for UAVs in real environment is complex. Hence, Without loss of generality, we create a virtual 3D environment with high matching degree to the real-world urban areas. Unlike most of the existing virtual environments, which are studied in literature and usually modeled as a grid world, in this paper, we focus on a free space environment containing 3D obstacles that may have diverse shapes as illustrated in Fig.~\ref{system}. Consequently, the UAV has the freedom to take any direction and speed to reach its target unlike grid world, which restricts the freedom of UAV into a finite set of actions. The goal is to train the UAV to fly safely from any arbitrary starting position to reach any destination in the considered area with continuous action space. The UAV, defined as $u$, is characterized by its 3D Cartesian geographical location $loc_u = [x, y, z]$ and initially situated at $loc_u(0) = [x_0, y_0, z_0]$. The destination $d$ is defined by its 3D location $loc_{d} = [x_{d}, y_{d}, z_{d}]$. In this paper, the investigated system assumes the following assumptions: 
\begin{itemize}
\item The environment obstacles have different heights. Each one of them is represented by a 3D polygon characterized by its the starting point [$x_{obs}$, $y_{obs}$], the set containing the edges of the base $edg_{obs}$, and its height $h_{obs}$. Hence, if having an altitude higher than the obstacle's height, the UAV can cross over the obstacles. Otherwise, the UAV can avoid it by flying around.
\item The destination location is known to the UAV and it can be either static or dynamic (i.e., the target location can evolve over time). If the destination location is dynamic then it follows a random pre-defined trajectory, that is unknown by the UAV. \vspace{-0.4cm}
\end{itemize}

\subsection{UAV Actions}
For each taken action, we assume that the UAV chooses a distance to cross according to a certain direction in the 3D space during $\Delta t$ units of time. The action is modeled using the spherical coordinates $(\rho, \phi, \psi)$ as follows:
\begin{align}
    &x_u =  x_u + \rho \sin{\phi} \cos{\psi}, \quad y_u =  y_u + \rho \sin{\phi},\notag\\
    &\text{and }z_u =  z_u + \rho \cos{\phi},
\end{align}
where $\rho$ is the traveled radial distance by the UAV in each step $\left(\rho \in [\rho_{min}, \rho_{max}]\right)$, where $\rho_{max}$ is the maximum distance that the UAV can cross during the step length $\Delta t$. Its value depends on the maximum speed of the UAV denoted by $v_{max}$. The parameter $\psi$ denotes the inclination angle $\left(\psi \in [0, 2\pi]\right)$, and $\phi$ represents the elevation angle $\left(\phi \in [0,\pi]\right)$. For instance:\\
$\bullet$ if $\rho = \rho_{max}$, $\phi = \pi$, and any value of $\psi$, the UAV moves by $\rho_{max}$ along the Z axis.\\
$\bullet$ if $\rho = \rho_{max}$, $\phi = \pi/2$, and $\psi=0$, the UAV moves along the x axis.\\
The distance between the UAV and its target is defined as $D(u,d)$. 
\begin{figure}[t!]
  \centerline{\includegraphics[width=8cm]{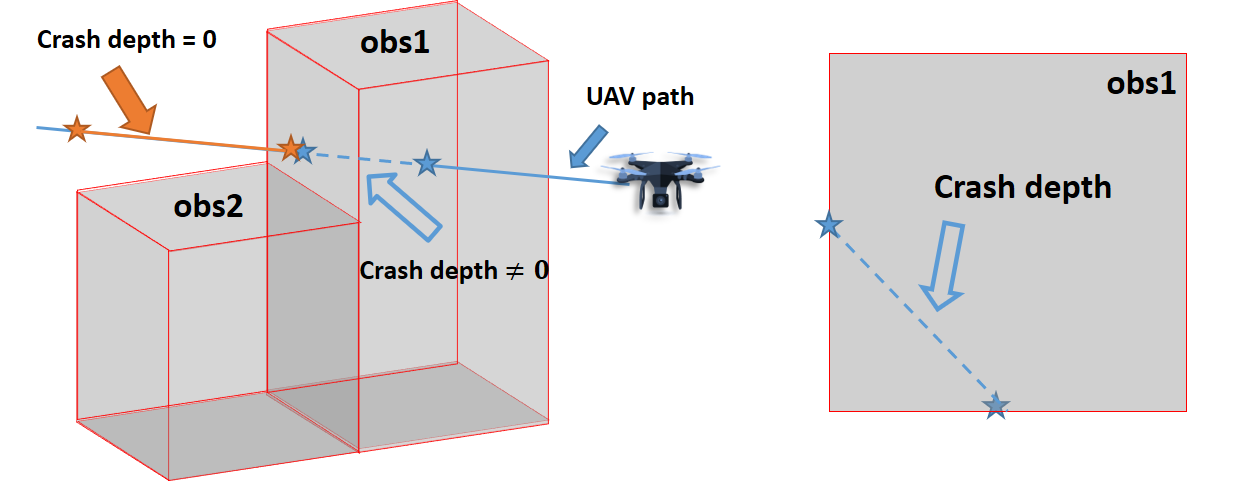}}\vspace{-0.1cm}
   \caption{Example representing a collision scenario. The UAV's altitude is less than the obstacle's height (obs1). The action is chosen such that the UAV crosses the obstacle. This results in an obstacle penalty reflecting the underwent depth.}\label{depth}\vspace{-0.4cm}
\end{figure}

\section{DDPG Learning for Autonomous Scheduling}\label{ProblemFormulation}
DDPG was developed as an extension of deep Q-network (DQN) algorithms introduced by Mnih et al.~\cite{mnih}, which was the first approach combining deep and reinforcement learning but only by handling low-dimensional action spaces. DDPG is also a deep RL algorithm, that has the capability to deal with large-dimensional/infinite action spaces. It tries to find an efficient behavior strategy for the agent to obtain maximal rewards in order to accomplish its assigned tasks~\cite{Lillicrap}. This DPG algorithm has the capability to operate over continuous action spaces which is a major hurdle for classic RL methods like Q-learning.

DDPG is based on the actor-critic algorithm. It is essentially a hybrid method that combines the policy gradient and the value function together. The policy function $\mu$ is known as the actor, while the value function $Q$ is referred to as the critic. Essentially, the actor output is an action chosen from a continuous action space, given the current state of the environment $a = \mu(s|\theta^{\mu})$, which, in our case, has the form of a tuple $a=[\rho,\phi,\psi]$. As for the critic, its output $Q(s, a|\theta^{\mu})$ is a signal having form of a Temporal Difference (TD) error to criticize the actions made by the actor knowing the current state of the environment. A diagram summarizing the actor-critic architecture is given in Fig.~\ref{alg_fig}. 

Note that the training phase of the DDPG model is executed for $M$ episodes where each one of them accounts for $T$ steps. We use the index $t$ to denote an iteration within a single episode where $t=1,\dots, T$. The actor and critic are designed with neural networks. The value network is updated based on Bellman equation~\cite{Murat} by minimizing the mean-squared loss between the updated Q value and the origin value, which can be formulated as shown in Algorithm 1 (line 11). As for the policy network's update (line 13), it is based on the deterministic policy gradient theorem~\cite{ddpg}. 

There are also some practical tricks that are used to enhance the performance of the framework. A trade off between exploration and exploitation is made by the use of $\epsilon$-greedy algorithm, where a random action $a_t$ is selecting with $\epsilon$ probability, otherwise a precise action $a_t=\mu(s_t|\theta^{\mu})$ is selected according to the current policy with a $1-\epsilon$ probability. Furthermore, an experience replay buffer $b$, with size $B$, is used during the training phase to break the temporal correlations. Each interaction with the environment is stored as tuples in the form of $[s_{t}, a, r, s_{t+1}]$, which are the current state, the action to take, the reward of performing action $a$ at state $s_t$, and the next state, respectively (Algorithm 1 (line 9)) and, during the learning phase, a randomly extracted set of data from the buffer is used (Algorithm 1 (line 10)). Also, target networks are exploited to avoid the divergence of the learning algorithm caused by the direct updates of the networks weights with the gradients obtained from the TD error signal. \vspace{-0.3cm}
\begin{algorithm}[t]
	\caption{DDPG}
	\label{alg:PP}
	\small
	\begin{algorithmic}[1]
		\STATE Randomly initialize critic $Q(s, a|\theta^{\mu})$ and actor 
		$\mu(s|\theta^{\mu})$  neural networks with weights $\theta^Q$ and $\theta^\mu$.
		\STATE Initialize target networks $Q'$ and $\mu'$ with weights $\theta^{Q'} \gets \theta^Q$, $\theta^{\mu'} \gets \theta^\mu $.
     	\STATE	Initialize replay buffer $b$.
        \FOR{ episode = 1,\dots,$M$}
		\STATE Receive first observation $s_1$.
		\FOR{ t = 1,\dots,$T$}
		\STATE Select $a_t$ based on $\epsilon$-greedy algorithm: select random action $a_t$ with $\epsilon$ probability, otherwise $a_t=\mu(s_t|\theta^{\mu})$  according to the current policy.
		\STATE Execute action $a_t$ and observe reward $r_t$ and new state $s_{t+1}$.
		\STATE Store transition $[s_t, a_t, r_t, s_{t+1}]$ in $b$.
		\STATE Sample a random batch of $N$ transitions $[s_j, a_j, r_j, s_{j+1}]$.
		\STATE Set $\hspace{0.1cm}y_j = r_j + \gamma Q'(s_{j+1}, \mu'(s_{j+1}|\theta^{\mu'} )|\theta^{Q'})$.
		\STATE Update critic by minimizing the loss: 
		
		$\hspace{1cm}L = \frac{1}{N}\sum_{j}(y_j - Q(s_j, a_j|\theta^Q))^2$
		\STATE Update the actor policy using policy gradient:
		
		\hspace{-0.5cm}$\nabla_{\theta^\mu}\mu|_{s_j}\approx \frac{1}{N}\sum_j\nabla_a Q(s, a|\theta^Q)|_{s=s_j,a=\mu(s_j)}\nabla_{\theta^{\mu}}\mu(s|\theta^\mu)|_{s_j} $
		\STATE Update the target networks: 
		
		$\hspace{1cm}\theta^{Q'} \gets \nu\theta^Q + (1-\nu)\theta^{Q'}$
		
		$\hspace{1cm}\theta^{\mu'} \gets \nu\theta^\mu + (1-\nu)\theta^{\mu'}$
		\ENDFOR
		\ENDFOR
	\end{algorithmic}
	\normalsize
\end{algorithm}
\setlength{\textfloatsep}{5pt}
\subsection{Reward Function}
In an obstacle-constrained environment, the UAV must avoid obstacles and autonomously navigate to reach its destination in real-time. Therefore, the reward function, denoted by $f_r$, is modeled such that it encourages the UAV to reach its destination and, at the same time, penalizes it when crashing. Thus, the reward function is composed of two terms: target guidance reward and obstacle penalty. The target guidance reward, denoted by $f_{gui}$,  is used to motivate the flying unit to reach its target as fast as possible, while the obstacle penalty, denoted by $f_{obp}\textbf{}$ is responsible for alerting the UAV to keep a certain safety distance off the obstacles. The reward function is formulated as follows:
\begin{subequations}
\begin{align}
&\hspace{-0.2cm}f_{r}(D(u,d), \sigma) = (1-\beta) f_{gui}(D(u,d))+\beta f_{obp}(\sigma),\\
&\hspace{-0.2cm}f_{gui}(D(u,d))  =  \exp(-5D(u,d)^2),\\
&\hspace{-0.2cm}f_{obp}(\sigma) = \exp(-100\sigma)-1, 
\end{align}
\end{subequations}
\begin{figure}[t]
  \centerline{\includegraphics[width=8cm]{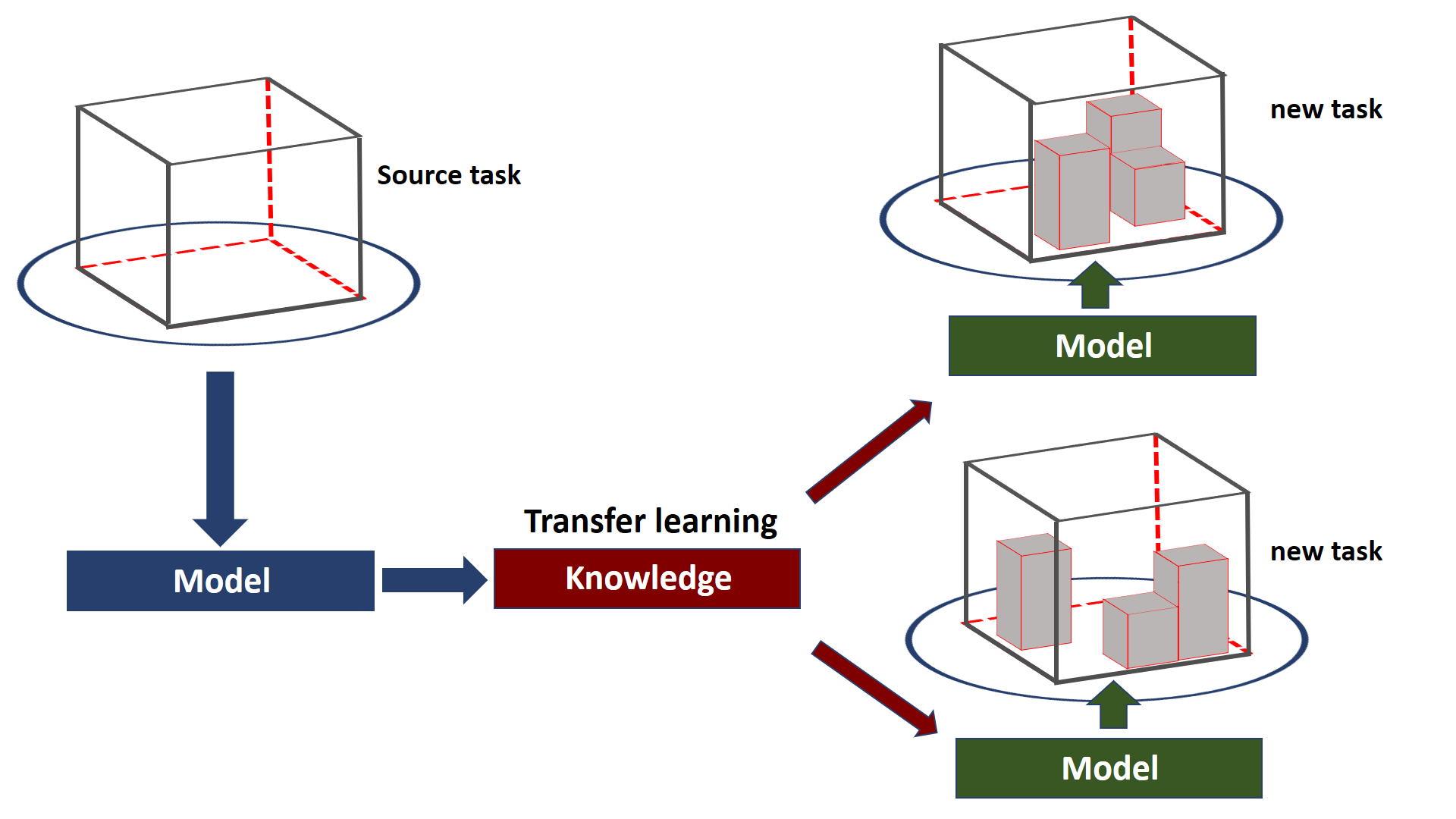}}
   \caption{Illustration of the transfer-learning technique.}\label{tramsfer}
\end{figure}
\begin{figure}[t]
  \centerline{\includegraphics[height=5cm,width=8cm]{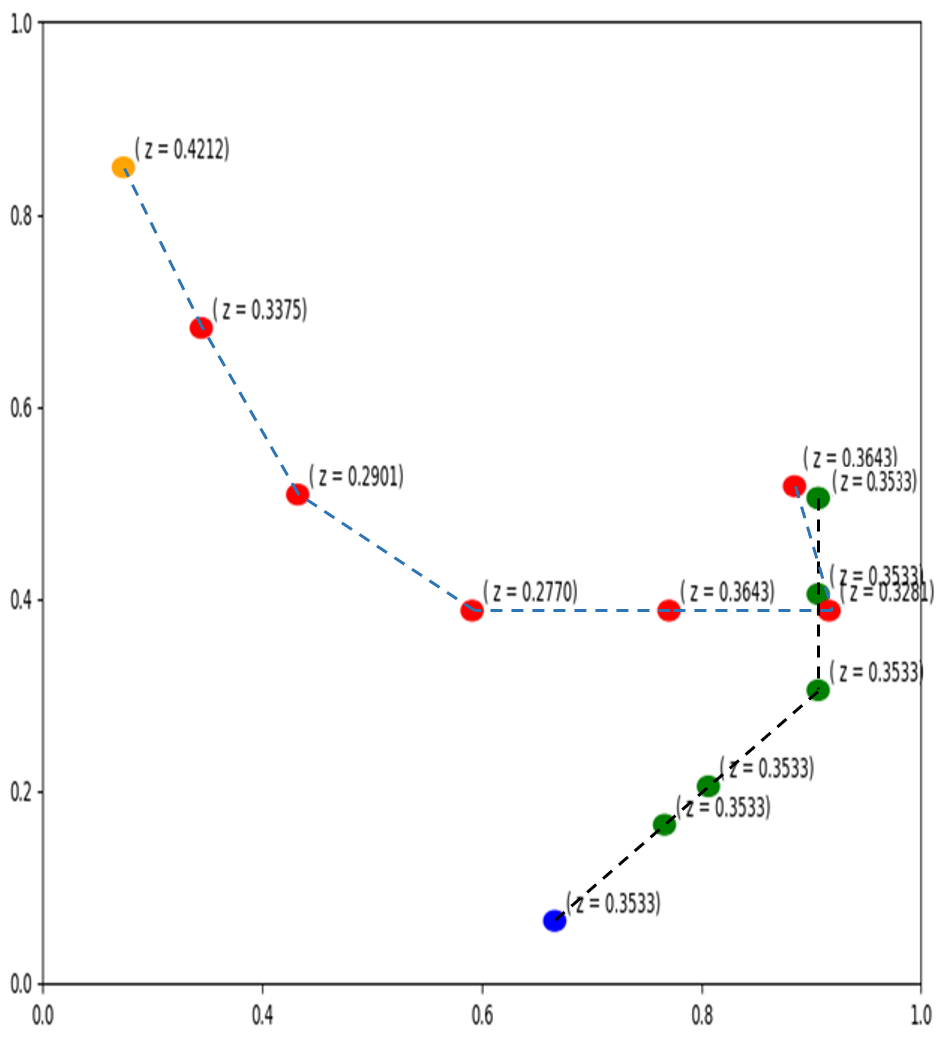}}\vspace{-0.2cm}
   \caption{Path followed by UAV based on the dynamic target location. Yellow dot: initial UAV position, blue dashed line: trajectory of the UAV, blue dot: target initial location, and dashed black line: the target trajectory.}\label{dyn}
\end{figure}
\begin{figure*}[t!]
  \centerline{\includegraphics[width=17cm, height= 5.75cm]{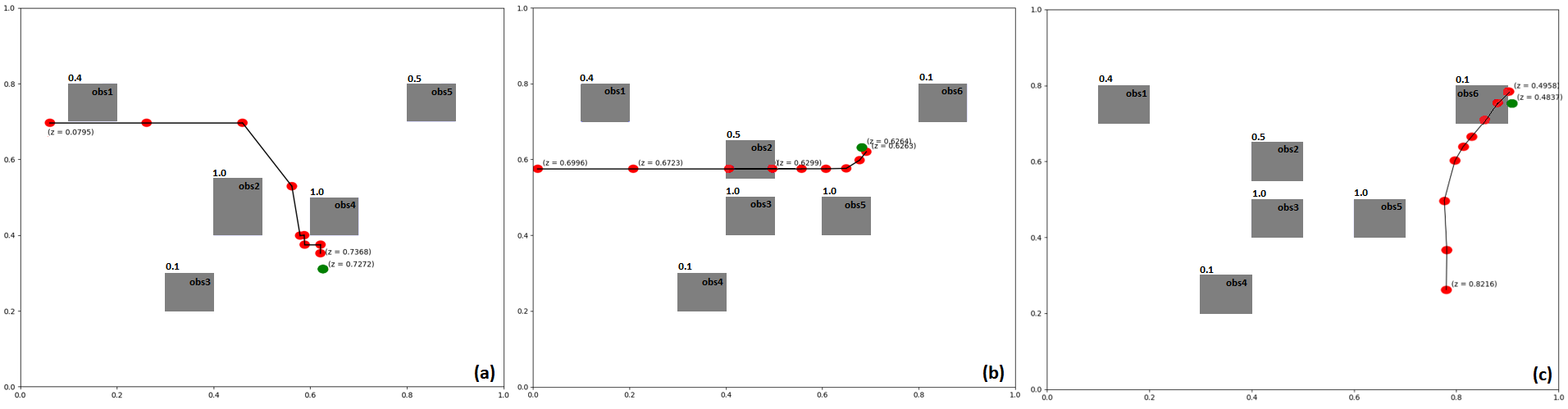}}\vspace{-0.2cm}
   \caption{Path followed by UAV based on the static target location. Red dots refers to the UAV, green dot refers to the target, and gray boxes to the obstacles. (a) environment 1 (env1) and (b, c) two cases in environment 2 (env2).}\label{sys}\vspace{-0.3cm}
\end{figure*}
where $\sigma$ is the crash depth explained in Fig.~\ref{depth} and $\beta$ is a variable that regulates the balance between $f_{obp}$ and $f_{gui}$. The obstacle penalty is modeled as a function of the crash depth $\sigma$ to conserve the continuous nature of the reward function instead of using discrete penalty, which proved to be more efficient to help the model to converge. In fact, when the crash depth is high, the UAV receives a higher penalty, whereas a small crash depth results in a lower penalty. The use of this approach helps the UAV learn efficiently over the training episodes how to adjust its trajectory to avoid obstacles.
\vspace{-0.3cm}
\subsection{Training Phase and Transfer Learning}
Transfer learning is a machine learning technique used to transfer the knowledge to speed up training and improve the performance of deep learning models. The proposed approach to train the UAV consists in two steps. Initially, we train the model in an obstacle-free environment. Training in such environment, grants the UAV the capability to reach any target in the covered 3D area with continuous space action. Then, the trained model on the obstacle-free environment will serve as a base for future models trained on other environments with obstacles. Afterwards, we transfer the acquired knowledge (i.e. source task) and use it to improve the UAV learning of new tasks where it updates its path based on the obstacle locations while flying toward its target. The adopted transfer learning technique applied to DDPG for autonomous UAV navigation is illustrated in Fig.~\ref{tramsfer}. Once the training phase is completed offline, the UAV is capable to make instant decisions, while interacting with the  environment, to manage real-time missions.
\section{Simulation Results}
In this section, we study the behavior of the system for selected scenarios. We also visualize the efficiency of the framework in terms of crash rate and tasks accomplishment. To do so, we assume that the UAV starting location $loc_u$, its target location $loc_{d}$, and the obstacles' parameters are randomly generated within a cube-shaped area with 100 $m$ edge length. We make sure that the locations of both the targets and the UAV are outside the obstacles. The rest of the simulation parameters are set as follows: 
\begin{table}[h]
\begin{center}  \vspace{-0.2cm}
\captionof{table}{Simulation parameters}
 \begin{tabular}{|c | c||c | c||c | c||c | c|}
 \hline

 $\beta$ & 4 & $\nu$ & 0.99 & $\rho_{\max}$ & 0.2 m & T & 100\\
 [1ex]\hline
  M & 40000 & $\epsilon_{end}$,$\epsilon_{start}$ & 0.1,0.9 & B & 10000 & N & 256\\
 \hline
\end{tabular}
\end{center}\vspace{-0.2cm}
\end{table}
 
\begin{figure}[h]
  \centerline{\includegraphics[width=9cm]{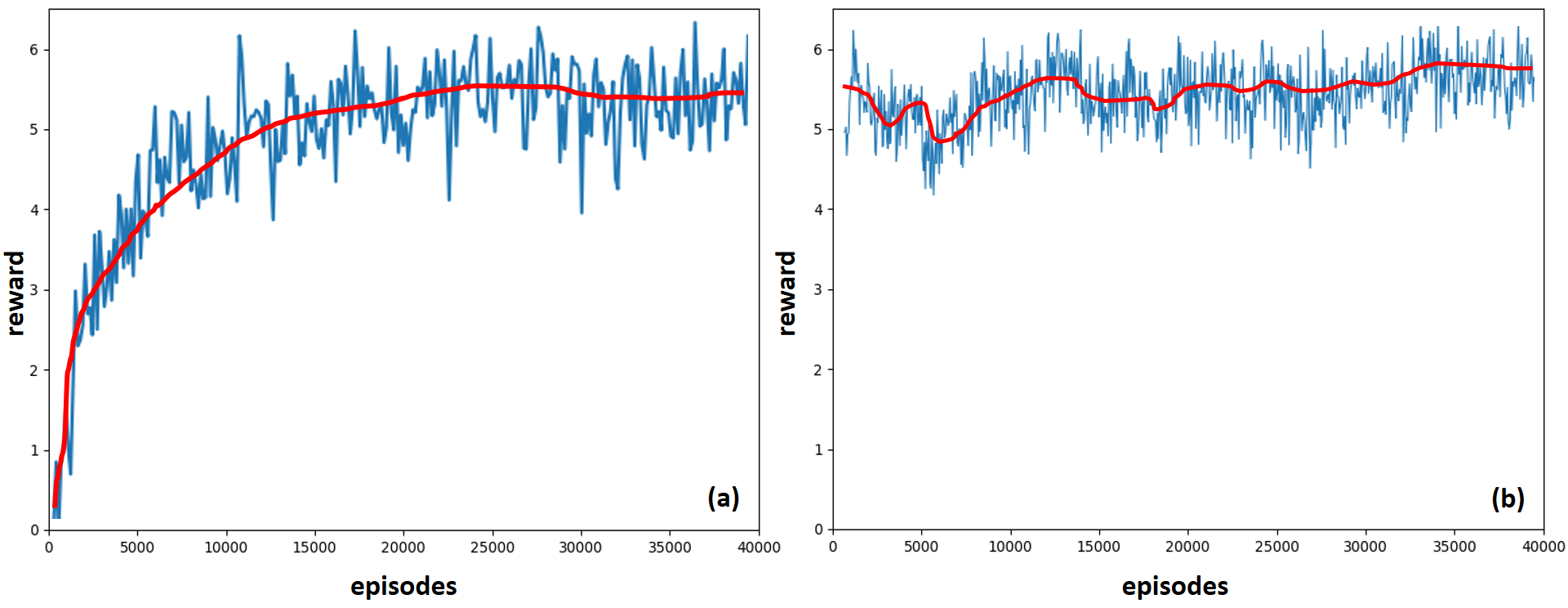}}\vspace{-0.2cm}
   \caption{The reward received by the UAV during the training phase: (a) the source task (obstacle-free) and (b) the environment with obstacles }\label{last}\vspace{-0.3cm}
\end{figure}

The simulations are executed using Python. For the sake of clarity, the figures concerning the UAV path planning are presented in only 2D dimension area (i.e., plan(x,y)) and we provide beside each dot, the altitude of either the target or the UAV. In the first scenario, we consider an obstacle-free environment. The destination location is assumed to be dynamic, that it keeps moving in a randomly generated way. As shown in Fig.~\ref{dyn}, the UAV is successfully adapting its trajectory based on the location of its target until it reaches it. This shows that the UAV succeeded in learning how to update each direction in order to ``catch'' its assigned destination. 

In the next scenarios, the obstacles are added in a random disposition with different heights as shown in Fig.~\ref{sys}. In these cases, we assume that the target destinations are static. In Fig.~\ref{sys}(a), the UAV successfully reached its destination location while avoiding the obstacles. In Fig.~\ref{sys}(b), on its way to the destination, the UAV crossed over $obs2$ ($z_u=0.63 > h_{obs2}= 0.5$) in order to reach faster its target location unlike the case in Fig.~\ref{sys}(a), where the UAV could not cross over $obs2$ to reach its destination as soon as possible because of the obstacle height (maximum height). In Fig.~\ref{sys}(c), having a higher altitude than $obs6$, the UAV crossed over $obs6$ to reach its target. In all cases, scenarios show some lacking in precision to reach the target location due to the fact of using infinite action space which makes it hard to get pinpoint accuracy. These scenarios showed that the UAV successfully learned how to avoid obstacles to reach its destination.

In Fig.~\ref{last}, we present the reward received by the UAV during its training phase, in Fig.~\ref{last}(a) shows that the UAV learns to obtain the maximum reward value in an obstacle-free environment. We successfully obtained a trained model capable of reaching targets in 3D environment with continuous action space. Then, using the knowledge gathered by the first training, we trained the model to be able to avoid obstacles. Fig.~\ref{last}(b) shows that the UAV model has converged and reached the maximum possible reward value.

During the testing phase and as shown in Table I, for the obstacle-free environment, the UAV successfully reached its target for the tested cases, 100\% success rate for 1000 test case. As for the environment with obstacles, in the case of env1, the UAV successfully reached its target safely for 84\% of the 1000 tested scenarios and in the case of env2, the reached its target safely for 82\% of the 1000 tested scenarios.
\begin{table}[t!]
\begin{center}  \vspace{-0.2cm}
\captionof{table}{Task-completion rate}
 \begin{tabular}{|c | c||c | c|}
 \hline
 
 Scenarios & obstacle free & obstacles env1 & obstacles env2  \\ [0.5ex] 
 \hline\hline
 Completion rate  & 100\%  & 84\%  & 82\%\\ [1ex] 

 \hline
\end{tabular}
\label{cc}
\end{center}\vspace{-0.2cm}
\end{table}
\section{Conclusion}
In this paper, we have developed an efficient framework for autonomous obstacle-aware UAV navigation in urban areas. Using a DDPG-based deep reinforcement learning approach, the UAV determines its trajectory to reach its assigned static or dynamic destination within a continuous action space. A transfer learning approach is devised in order to maximize a reward function balancing between target guidance and obstacle penalty. The simulation results exhibit the capability of UAVs in learning from the surrounding environment to determine their trajectories in real-time. %In our ongoing work, we will explore the challenges of operating a fleet of UAVs in more complex environment.

%\section{Conclusion}

\FloatBarrier
\bibliographystyle{IEEEtran}
%\bibliography{ISCAS}

\end{document}